\documentclass[10pt,journal]{IEEEtranTCOM}

\usepackage[english]{babel}
\usepackage[usenames]{color}
\usepackage[cp1250]{inputenc}
\usepackage{amsfonts}
\usepackage{amsthm}
\usepackage{graphicx}
\usepackage{epsfig}
\usepackage{mathrsfs}
\usepackage{amsmath}
\usepackage{algorithm}
\usepackage{algorithmic}
\usepackage{hyperref}
\DeclareMathOperator*{\argmax}{arg\,max}
\DeclareMathOperator*{\argmin}{arg\,min}

\theoremstyle{plain}

\begin{document}
\newcommand{\bea}{\begin{eqnarray}}
\newcommand{\eea}{\end{eqnarray}}
\newcommand{\be}{\begin{equation}}
\newcommand{\ee}{\end{equation}}
\newcommand{\beas}{\begin{eqnarray*}}
\newcommand{\eeas}{\end{eqnarray*}}
\newcommand{\bs}{\backslash}
\newcommand{\bc}{\begin{center}}
\newcommand{\ec}{\end{center}}
\def\SC {\mathscr{C}}

\title{Adaptive exponential power distribution\\with moving estimator\\ for nonstationary  time series}
\author{\IEEEauthorblockN{Jarek Duda}\\
\IEEEauthorblockA{Jagiellonian University,
Golebia 24, 31-007 Krakow, Poland,
Email: \emph{dudajar@gmail.com}}}
\maketitle

\begin{abstract}
While standard estimation assumes that all datapoints are from probability distribution of the same fixed parameters $\theta$, we will focus on maximum likelihood (ML) adaptive estimation for nonstationary time series: separately estimating parameters $\theta_T$ for each time $T$ based on the earlier values $(x_t)_{t<T}$ using (exponential) moving ML estimator $\theta_T=\argmax_\theta l_T$ for  $l_T=\sum_{t<T} \eta^{T-t} \ln(\rho_\theta (x_t))$ and some $\eta\in(0,1]$. Computational cost of such moving estimator is generally much higher as we need to optimize log-likelihood multiple times, however, in many cases it can be made inexpensive thanks to dependencies. We focus on such example: $\rho(x)\propto \exp(-|(x-\mu)/\sigma|^\kappa/\kappa)$  exponential power distribution (EPD) family, which covers wide range of tail behavior like Gaussian ($\kappa=2$) or Laplace ($\kappa=1$) distribution. It is also convenient for such adaptive estimation of scale parameter $\sigma$ as its standard ML estimation is $\sigma^\kappa$ being average $\|x-\mu\|^\kappa$. By just replacing average with exponential moving average: $(\sigma_{T+1})^\kappa=\eta(\sigma_T)^\kappa +(1-\eta)|x_T-\mu|^\kappa$ we can inexpensively make it adaptive. It is tested on daily log-return series for DJIA companies, leading to essentially better log-likelihoods than standard (static) estimation, with optimal $\kappa$ tails types varying between companies. Presented general alternative estimation philosophy provides tools which might be useful for building better models for analysis of nonstationary time-series.
\end{abstract}
\section{Introduction}
In standard parametric estimation we choose some density family $\rho_{\theta}$ and assume that all datapoints are from this distribution using the same parameters $\theta$. For maximum likelihood (ML) estimation we find $\theta$ maximizing $l=\sum_{t=1}^n \frac{1}{n}\ln(\rho_{\theta}(x_t))$ having equal $1/n$ contribution of all $n$ datapoints $(x_t)_{t=1..n}$. This estimation is perfect for i.i.d. sequence from stationary time series. For distinction, in analogy to static-adaptive separation of models in data compression~\cite{adap}, let us refer to it as \textbf{static estimation}.

\begin{figure}[t!]
    \centering
        \includegraphics{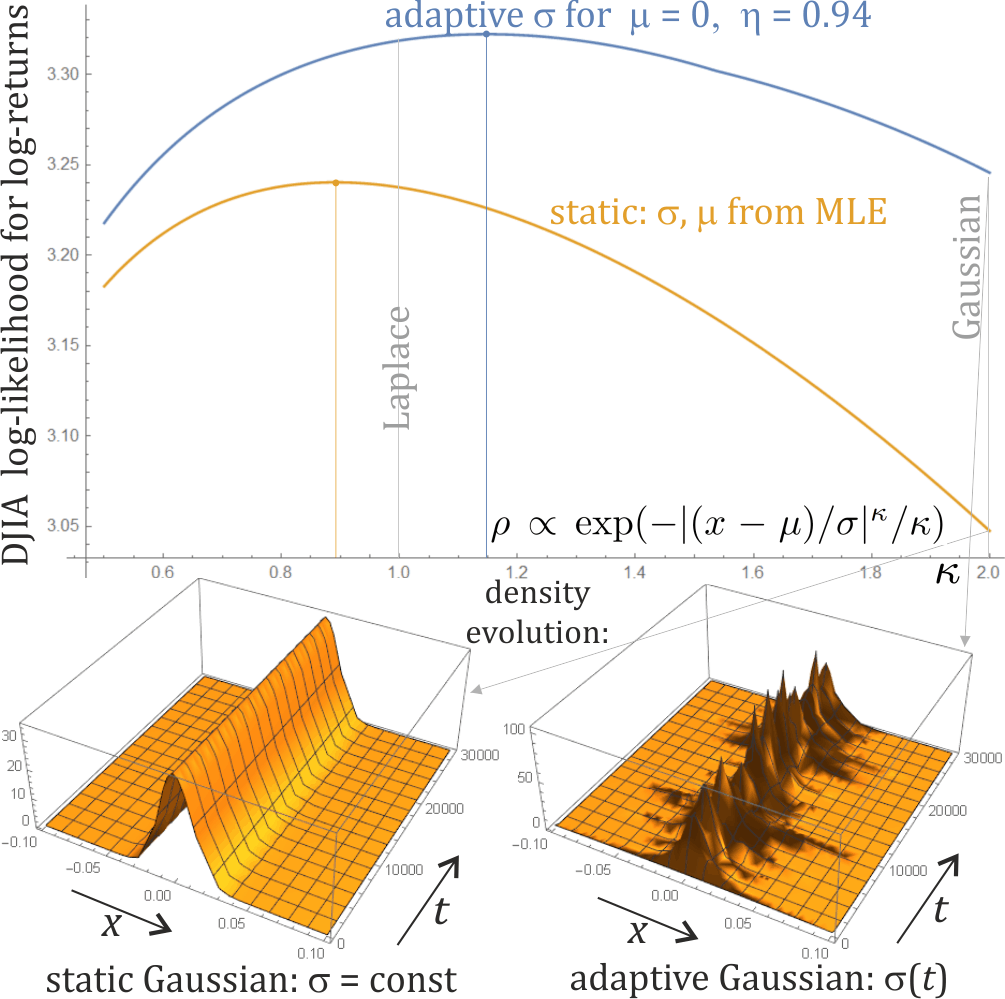}
        \caption{Log-likelihood $l=\frac{1}{n}\sum_T \ln(\rho_{\theta_T}(x_T))$ evaluation for 100 years daily log-returns of DJIA sequence. In horizontal axis there is shape parameter $\kappa$ of $\rho(x)\propto \exp(-|(x-\mu)/\sigma|^\kappa/\kappa)$ exponential power distribution. Orange line shows evaluation of standard "static" model $\theta_T=\theta$: MLE choosing fixed $\sigma,\mu$ parameters, separately for each $\kappa$. Blue line shows evaluation of the simplest adaptive model: separately for each $\kappa$, fixing $\mu=0$, evolving scale parameter $\sigma^\kappa$ as exponential moving average of $|x-\mu|^\kappa$ up to the previous position: $(\sigma_{T+1})^\kappa=\eta(\sigma_T)^\kappa +(1-\eta)|x_T-\mu|^\kappa$. We can see that 1) adaptivity brings large log-likelihood improvements, 2) the optimal $\kappa$ (marked with dots) is far from Gaussian, much closer to Laplace distribution (heavier tails), 3) optimal $\kappa$ for static and adaptive models are different.}
       \label{djia}
\end{figure}

For nonstationary time series these $\theta$ parameters might evolve in time, like estimated density in the bottom of Fig. \ref{djia}. To estimate such parameter evolution, we focus here on \textbf{adaptive estimation} using moving estimator~\cite{adapt}, which for time $T$ finds $\theta_T$ maximizing moving likelihood $l_T$ based only on the previously seen datapoints, for example using exponentially weakening weights:
\be \theta_T=\argmax_\theta l_T\qquad l_T = \sum_{t<T} \eta^{T-t} \ln(\rho_{\theta}(x_t))\ee
where $\eta\in(0,1]$ defines rate of weakening of contribution of the old points in such exponential moving average. For $\eta=1$ it becomes ML estimation based on all previous points. In practice usually $\eta\in (0.9,1)$, generally might differ between parameters (e.g. here $\eta\approx 0.94$ for $\sigma$, $\nu\approx 0.997$ for $\mu$).

\begin{figure*}[h!]
    \centering
        \includegraphics{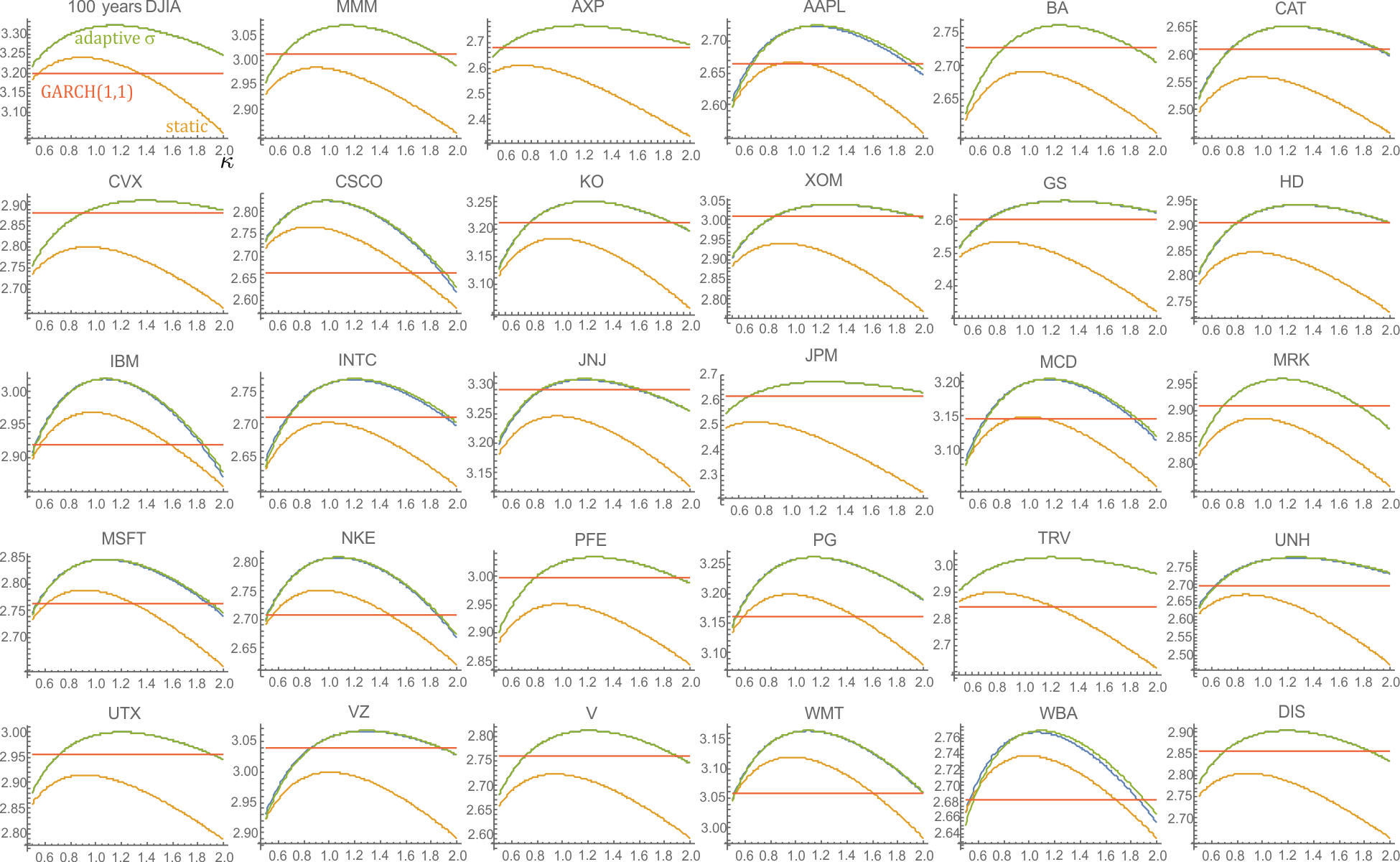}
        \caption{Analogously as in Fig. \ref{djia}, log-likelihood (vertical) dependence from shape parameter $\kappa$ (horizontal) for modeling of daily log-returns for 100 years DJIA, and 10 years for 29 of its recent companies. The lowest orange plots are for standard static MLE estimation of EPD. The higher green plots are for discussed EPD adaptive $(\sigma_{T+1})^\kappa=\eta(\sigma_T)^\kappa +(1-\eta)|x_T-\mu|^\kappa$ estimation using $\eta=0.94$ rate. The blue lines use individually optimized $\eta$ rate instead - usually is nearly the same, log-likelihood improvement from its optimization is nearly negligible. For comparison, there is also plotted red line for evaluation using standard GARCH(1,1) model - it uses Gaussian distribution corresponding to $\kappa=2$, where we can see it is comparable to adaptive EPD. While there is usually assumed $\kappa=2$ type of tail behavior, we can see that data suggest much lower optimal $\kappa$, closer to $\kappa=1$ of Laplace distribution. Moreover, while intuition suggests some universality of tail behavior, data shows that it varies between companies. Additionally, we can see that optimal $\kappa$ is larger for adaptive estimation - intuitively, adaptivity has allowed to use thinner tails.}
       \label{stock}
\end{figure*}

While standard static estimation is performed once - finding a compromise for all datapoints, discussed adaptive estimation is generally much more computationally expensive - needs to be performed separately for each $T$. However, in some situations it can be optimized at least for some parameters, by making it an evolving estimation exploiting previously found state. For example when standard ML estimation is given by average over some function of datapoints, we can transform it to adaptive estimation by just replacing this average with exponential moving average.

Specifically, we will focus on exponential power distribution (EPD)~\cite{epd} family: $\rho(x)\propto \exp(-|(x-\mu)/\sigma|^\kappa/\kappa)$, which covers wide range of tail behaviors like Gaussian ($\kappa=2$) or Laplace ($\kappa=1$) distribution. It is also convenient for such adaptive estimation of scale parameter $\sigma$ as in standard ML estimation: $\sigma^\kappa$ is average of $|x_t-\mu|^\kappa$. We can transform it to adaptive estimation by just replacing average with exponential moving average: $(\sigma_{T+1})^\kappa=\eta(\sigma_T)^\kappa +(1-\eta)|x_T-\mu|^\kappa$.

On example of 100 years Dow Jones Industrial Average (DJIA) daily log-returns and 10 years for 29 its recent companies, we have tested that such adaptive estimation of $\sigma$ leads to essentially better log-likelihoods than standard static estimation as we can see in Fig. \ref{djia}, \ref{stock}. Surprisingly, the $\kappa$ parameter defining tail behavior, usually just chosen as $\kappa=2$ by assuming Gaussian distribution, turns out less universal - various companies have different optimal $\kappa$, much closer to heavier tail $\kappa=1$ of Laplace distribution.

The discussed general philosophy of adaptive estimation directly focuses on non-stationarity of time series - trying to model evolution of parameters. Its applications like  adaptive EPD can be used as a building block for the proper methods like ARIMA-GARCH family. Surprisingly, such  adaptive EPD $\sigma$ estimation (just $(\sigma_{T+1})^\kappa=\eta(\sigma_T)^\kappa +(1-\eta)|x_T-\mu|^\kappa$) for this data already turns out comparable with much more sophisticated standard methods like GARCH(1,1)~\cite{garch}, represented as red lines in Fig. \ref{stock}. These more sophisticated models assume some arbitrary evolution of parameters, while moving estimator does not do it (is agnostic) - just shifts the estimator to get local parameters.

\section{Exponential power distribution (EPD)}
\begin{figure}[t!]
    \centering
        \includegraphics{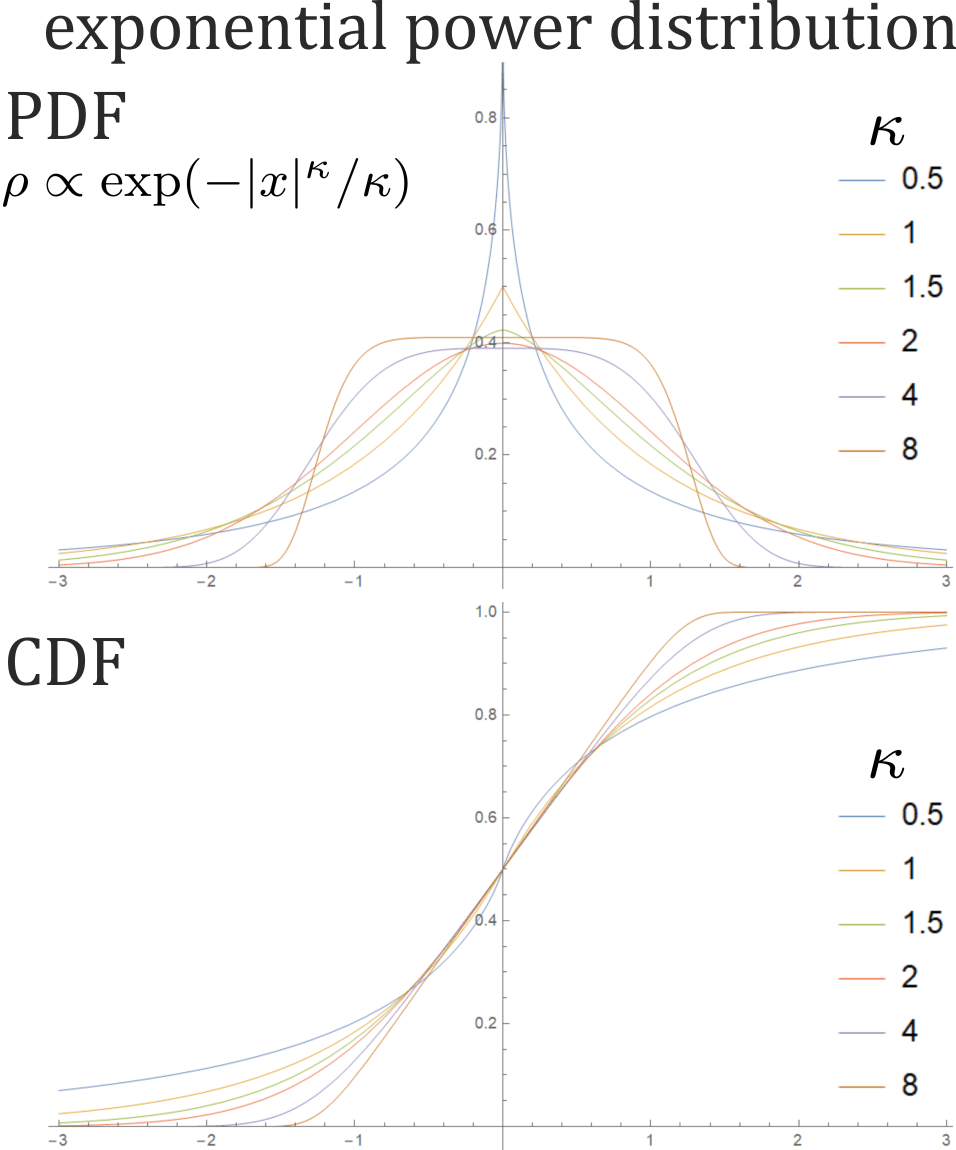}
        \caption{Probability distribution function (PDF, $\rho\propto \exp(-|x|^\kappa/\kappa)$) and cumulative distribution function (CDF) for exponential power distribution (EPD) with fixed center $\mu=0$ and scale parameter $\sigma=1$, for various shape parameters $\kappa$. We get Gaussian distribution for $\kappa=2$, Laplace distribution for $\kappa=1$, and can also cover different types of tails and bodies of distribution, choosing $\kappa$ agnostically: from evaluation on data. We could also use asymmetric EPD~\cite{aepd}, e.g. by using separate  $\kappa$ for each direction.}
       \label{EPD}
\end{figure}

For $\kappa>0$ shape parameter, $\sigma>0$ scale parameter and $\mu\in\mathbb{R}$ location, probability distribution function (PDF, $\rho_{\kappa\mu \sigma}$) and cumulative distribution function (CDF, $F_{\kappa\mu \sigma}(x)=\int_{-\infty}^x \rho_{\kappa\mu \sigma}(y)dy$) of EPD are correspondingly:

\be \rho_{\kappa\mu \sigma}(x)=\frac{1}{2\sigma}\frac{\kappa^{-1/\kappa}}{ \Gamma(1+1/\kappa)} e^{-\frac{1}{\kappa}\left(\frac{|x-\mu|}{\sigma}\right)^\kappa} \label{EPDrho}\ee
$$F_{\kappa\mu \sigma}(x) = \begin{cases}
      \frac{1}{2}\gamma\left(\frac{1}{\kappa},\frac{(|x-\mu|/\sigma)^\kappa}{\kappa}\right) & \text{if}\ x<\mu \\
      1-\frac{1}{2}\gamma\left(\frac{1}{\kappa},\frac{(|x-\mu|/\sigma)^\kappa}{\kappa} \right) & \text{if}\ x\geq \mu
    \end{cases}$$

where $\Gamma$ is Euler gamma function, $\gamma(a,z)=\Gamma(a,z)/\Gamma(a)$ is regularized incomplete gamma function. These PDF and CDF are visualized in Fig. \ref{EPD}.
\subsection{Static parameter estimation}
Let us start with standard static ML estimation: assuming $\{x_i\}_{i=1..n}$ i.i.d sequence. For generality let use weights $w_i$ of points, assuming $\sum_i w_i=1$ to imagine them as contribution of each point. In standard static estimation we assume equal $w_i=1/n$ contributions.

Such general weighted log-likelihood is:
\be l=\sum_{i=1}^n  w_i\,\ln\left(\rho_{\kappa\mu \sigma}(x_i)\right)=\label{ll}\ee
$$-\ln\left(2\sigma\kappa^{1/\kappa}\Gamma\left(1+1/\kappa\right)\right)
-\frac{1}{\kappa \sigma^\kappa}\sum_{i=1}^n w_i\,|x_i-\mu|^\kappa$$
From $\partial l / \partial \sigma = 0$ necessary condition we get maximum likelihood estimator for scale parameter (assuming fixed $\kappa, \mu$):
$$0=\frac{\partial l}{\partial \sigma}= -\frac{1}{\sigma}+\frac{1}{\sigma^{\kappa+1}} \sum_{i=1}^n w_i\,|x_i-\mu|^\kappa$$
\be \hat{\sigma}=\argmax_\sigma \, l =\left(\sum_{i=1}^n w_i\, |x_i-\mu|^\kappa\right)^{1/\kappa} \label{best} \ee

There is no general analytic formula for the remaining parameters, but they can be estimated numerically. Estimation of the location $\mu$ can be expressed as:
\be \hat{\mu}=\argmin_\mu\, \sum_{i=1}^n w_i\, |x_i-\mu|^\kappa  \ee
for Gaussian distribution $(\kappa=2)$ it is just mean of values $\hat{\mu}=\sum_i w_i x_i$. For Laplace distribution $(\kappa=1)$ it is their median. Some practical approximation, e.g. as initial value of more sophisticated estimation, might be just using mean for all $\kappa$.

To approximately estimate the shape parameter $\kappa$, we can for example use the  method of moments, especially that variance of EPD has simple form:
\be \textrm{variance} = \frac{\kappa^{2/\kappa}\, \Gamma(3/\kappa)}{\Gamma(1/\kappa)}\, \sigma^2 \label{variance} \ee
which is strongly decreasing with $\kappa$, e.g. $2\sigma^2$ for Laplace distribution, $\sigma^2$ for Gaussian distribution.

\subsection{Adaptive estimation of scale parameter $\sigma$}
Let us define moving log-likelihood for time $T$ using only its previous points, exponentially weakening weights of the old points to try to estimate parameters describing local behavior:
\be l_{T}=  \sum_{t=1}^{T-1} w_{T,t}\, \ln(\rho_{\kappa\mu \sigma}(x_t))\ee
$$\textrm{for}\quad w_{T,t} = \frac{\eta^{T-t}}{c_T}\qquad c_T=\sum_{t=1}^{T-1}\eta^{T-t}=\frac{\eta-\eta^T}{1-\eta}$$
to get (exponential moving) weights summing to 1.

Now fixing $\kappa$, $\mu$ and optimizing $\sigma$, from (\ref{best}) we get
\be \hat{\sigma}_T=\argmax_\sigma \, l_T =\left(b_T\right)^{1/\kappa} \ee
$$ \textrm{for}\qquad (\hat{\sigma}_T)^\kappa=b_T = \sum_{t=1}^{T-1} \frac{\eta^{T-t}}{c_T}\, |x_t-\mu|^\kappa $$
which is exponential moving average (EMA), can be evolved iteratively:
\be b_{T+1} = \eta\, b_T + (1-\eta)\, |x_T-\mu|^\kappa \ee
initial $b_1=(\hat{\sigma}_1)^\kappa$ has to be chosen arbitrarily.

We have transformed static estimator given by average, into adaptive estimator by just replacing average with  exponential moving average. Observe that it can be analogously done for any estimator of $\hat{\theta} = f(\sum_i w_i g(x_i))$ form.



\subsection{Generalization, interpretation and choice of rate $\eta$}
To generalize the above, assume that estimation of parameter $\theta$ is analogously given by average ($\sum_i w_i=1$):
\be\hat{\theta}=f\left(\sum_i w_i g(x_i)\right) \ee
For the above $\sigma$ parameter of EPD with $\kappa,\mu$ fixed, we would have $f(x)=x^{1/\kappa}$, $g(x)=|x-\mu|^\kappa$.

Generally we analogously have $\hat{\theta}_T=f(b_T)$ adaptation for
\be b_{T+1} = \eta\, b_T + (1-\eta)\, g(x_T) \ee
Denoting $\bar{\eta}=1-\eta$, we can write it as:
\be b_{T+1}=b_T +\bar{\eta}\, (g(x_T)-b_T) \label{rw}\ee
allowing to imagine evolution of $b$ as random walk with step from $\bar{\eta}\, (g(X)-b)$ random variable, which can evolve in time here. Generally $\bar{\eta}=1-\eta$ is proportional to speed of this random walk.\\

This interpretation could be used to optimize the choice of $\eta$, separately for each parameter, also its potential evolution. For example by calculating (e.g. exponential moving averaged) square root of variance of $(b_{T+1}-b_T)_T$ sequence, and evaluate square root of variance of $g(X)-b$ random variable - dividing them we get estimation of $\bar{\eta}=1-\eta$ parameter. \\

We can also try to adapt $\eta$ parameter based on data to optimize some final evaluation like:
$$l=\frac{1}{n}\sum_{T=1}^n \hat{l}(\theta_T, x_T)\qquad \textrm{e.g. for}\quad \hat{l}(\theta, x)=\ln(\rho_\theta(x))$$
for log-likelihood, or e.g. minus squared error for MSE.

E.g. using \ref{rw} recurrence, we can condition its time $T$ term with the current $\bar{\eta}=1-\eta$ rate, $\theta_T=f(b_T)$:
\be\frac{\partial}{\partial_{\bar{\eta}}} \hat{l}\left(\theta_T, x_T\right)=\frac{\partial}{\partial_{\bar{\eta}}} \hat{l}\left(f(b_T), x_T\right)=\ee
$$=\frac{\partial}{\partial_{\bar{\eta}}} \hat{l}(f(b_{T-1}+\bar{\eta}(g(x_{T-1})-b_{T-1})), x_T)\approx$$
$$\approx (g(x_{T-1})-b_{T-1})\,f'(b_{T-1})\, (\partial_\theta \hat{l})(f(b_{T-1}),x_T):=G_T $$
what allows e.g. for gradient optimization of $\eta$ for the next step, like for some tiny $\epsilon>0$ use $\eta_{T+1}= \eta_T -\epsilon G_T$ update.\\

We leave its details for future work as improvement by optimization from the fixed $\eta=0.94$ was practically negligible for the analyzed daily log-return data. Fig. \ref{stock} presents such difference by green plot for $\eta=0.94$, and blue for individually optimized $\eta$.

\subsection{Approximated adaptive estimation of $\mu$, $\kappa$}
While in practice the most important seems adaptation of scale parameter $\sigma$, there might be also worth to consider adaptation of the remaining parameters. Their estimation  rather does not have analytical formulas already in static case, hence for adaptive case we should look for practical approximations, preferably also based on EMA for inexpensive  updating.\\

Location $\mu$ is mean of such parametric distribution, just using mean of datapoints as its estimator is optimal for Gaussian   distribution case $(\kappa=2)$, and can be easily transformed to adaptive estimation. Hence a natural approximated estimation is analogous:
\be \hat{\mu}_{T+1}= (1-\nu) \hat{\mu}_{T} +\nu\, x_T\ee
for e.g. $\hat{\mu}_1 = 1$ and some chosen rate $\nu$, not necessarily equal $\eta$ (for this data $\eta\approx 0.94$, $\nu\approx 0.997$).\\

Adaptive estimation of $\kappa$ seems more difficult. Some example of approximation is using method of moments e.g. with (\ref{variance}) formula, especially that we can naturally get adaptive estimation of moments with EMA. For example as $\textrm{variance}_T = \widehat{x^2}_T - (\hat{\mu}_T)^2$ here using additional analogous EMA for estimated recent mean $x^2$:
$$ \widehat{x^2}_{T+1}= (1-\nu)\, \widehat{x^2}_{T} +\nu (x_T)^2. $$
Another general approach are gradient methods, adapting chosen parameter(s) e.g. to increase log-likelihood contribution. For example for some tiny $\epsilon>0$:
$$\kappa_{T+1}=\kappa_T + \epsilon \frac{\partial}{\partial \kappa} \ln(\rho_{\kappa\mu\sigma}(x_T))$$

\section{DJIA log-returns tests}
We will now look at evaluation of these methodologies from perspective of $\approx 100$ years daily Dow Jones index\footnote{Source of DJIA time series: http://www.idvbook.com/teaching-aid/data-sets/the-dow-jones-industrial-average-data-set/},  values $v$, working on $x_t=\ln(v_{t+1})-\ln(v_t)$ log-returns sequence for $t=1,\ldots, n$ for $n=29354$, summarized in Fig. \ref{djia}.

As evaluation there is used mean log-likelihood: $l=\frac{1}{n} \sum_{t=1}^n \ln(\rho_t(x_t))$, where in static setting $\rho_t$ has constant parameters chosen by MLE, in adaptive these parameters evolve in time: are estimated based on previous values.
\begin{figure}[t!]
    \centering
        \includegraphics{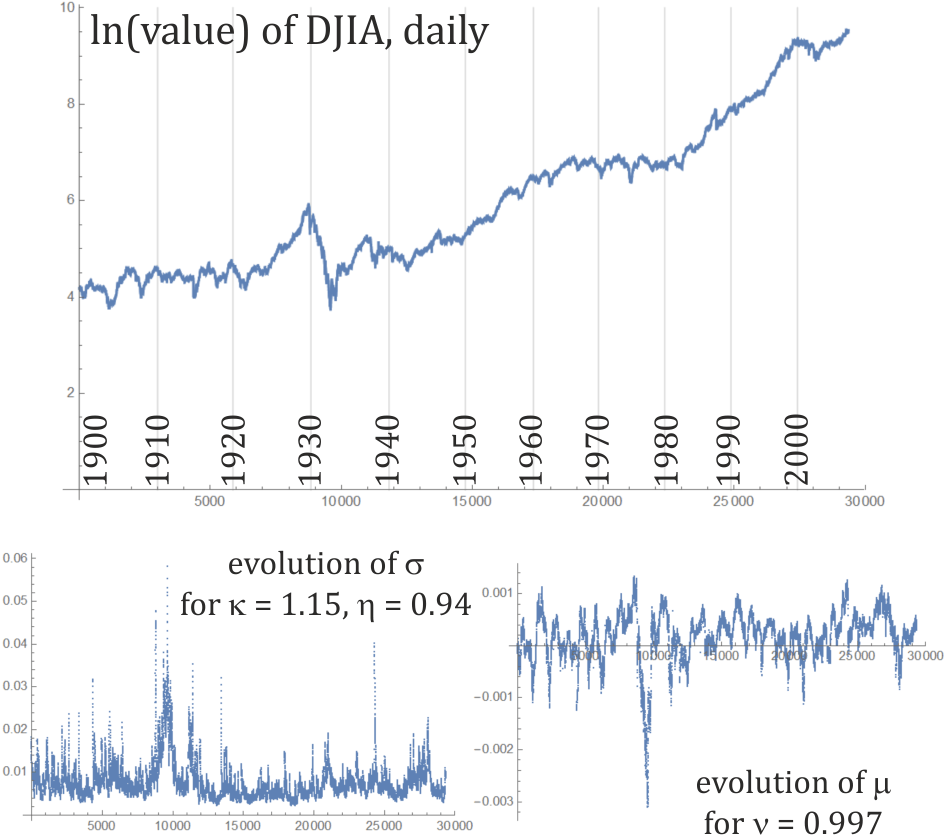}
        \caption{Top: log-values of DJIA and corresponding dates. Bottom:
        obtained evolution of $\sigma$ and $\mu$ parameters for DJIA log-returns adaptive models (EMA). Maxima of the former correspond to locally increased variance, maxima/minima of the latter correspond to periods of ascend/descend.}
       \label{bmu}
\end{figure}

In adaptive settings there was arbitrarily chosen initial $\sigma_1=0.01$, $\mu_1=0$, and from numerical search: $\eta=0.94$, $\nu=0.997$. Here are the obtained parameters and mean log-likelihoods for various settings:

\begin{itemize}
  \item static Gaussian ($\kappa=2$) distribution has MLE mean $\mu\approx 0.00018$, $\sigma\approx 0.0115$, giving $l\approx 3.04756$,
  \item static Laplace ($\kappa=1$) distribution has MLE median $\mu\approx 0.00044$, $\sigma\approx 0.00722$, giving $l\approx 3.23749$,
  \item static EPD has MLE $\kappa\approx 0.8912$, $\mu\approx 0.00046$, $\sigma\approx 0.00686$, giving $l\approx 3.2403$,
  \item Gaussian with $\mu=0$ and adaptive $\sigma$ gives $l\approx 3.2456$,
  \item Laplace with $\mu=0$ and adaptive $\sigma$ gives $l\approx 3.3187$,
  \item EPD optimal $\kappa\approx 1.1472$ with $\mu=0$ and adaptive $\sigma$ gives $l\approx 3.3222$,
  \item Gaussian with adaptive $\sigma$ and $\mu$ gives $l\approx 3.2452$,
  \item Laplace with adaptive $\sigma$ and $\mu$ gives $l\approx 3.3207$,
  \item EPD $\kappa=1.15$ with adaptive $\sigma$ and $\mu$ gives $l\approx 3.3234$.
\end{itemize}
We can see that standard assumption of static Gaussian can be essentially improved both by going to closer to Laplace distribution, and by switching to adaptive estimation of scale parameter $b$. Additional adaptive estimation of $\mu$ location provides some tiny further improvement here. The final used evolution of $\sigma$ and $\mu$ is presented in Fig. \ref{bmu}.

There were also trials for adapting $\kappa$, but were not able to provide a noticeable improvement here.

Figure \ref{stock} additionally contains such evaluation of log-returns for 29 out of 30 companies used for this index in September 2018. Daily prices for the last 10 years were downloaded from NASDAQ webpage (www.nasdaq.com) for all but DowDuPont (DWDP) - there were used daily close values for 2008-08-14 to 2018-08-14 period ($2518$ values) for the remaining 29 companies: 3M (MMM), American Express (AXP), Apple (AAPL), Boeing (BA), Caterpillar (CAT), Chevron (CVX), Cisco Systems (CSCO), Coca-Cola (KO), ExxonMobil (XOM), Goldman Sachs (GS), The Home Depot (HD), IBM (IBM), Intel (INTC), Johnson\&Johnson (JNJ), JPMorgan Chase (JPM), McDonald's (MCD), Merck\&Company (MRK), Microsoft (MSFT), Nike (NKE), Pfizer (PFE), Procter\&Gampble (PG), Travelers (TRV), UnitedHealth Group (UNH), United Technologies (UTX), Verizon (VZ), Visa (V), Walmart (WMT), Walgreens Boots Alliance (WBA) and Walt Disney (DIS).

\subsection{Further improvements with Hierarchical Correlation Reconstruction}\label{HCR}
The estimated parametric distributions of variables in separate times often leave statistical dependencies which can be further exploited.

For this purpose, we can use the best found model, here EPD $\kappa=1.15$ with adaptive $\sigma$ and $\mu$ for DJIA sequence, and use it for normalization of variables to nearly uniform distributions by going through cumulative distribution functions (CDF) of estimated parametric distributions: transform to $\{y_T\}_T =\{\textrm{CDF}_T(x_T)\}_{T=1..n}$ sequence.

We can then take e.g. $d$ neighboring values of $\{y\}$ sequence, which should be from approximately  uniform distribution on $[0,1]^d$. In Hierarchical Correlation Reconstruction~\cite{hcr} we estimate distortion from this uniform distribution as a polynomial of modelled static or adaptive coefficients.

Obtained mean log-likelihood improvement for $d=1$ single variables was $\approx 0.0058$ for static model (in 10-fold cross-validation), $\approx 0.0072$ for adaptive (MSE moving estimator) - using polynomial model to improve the original EPD model. Polynomial model can improve behavior of body of the distribution, but has not much influence on the tails - EPD should mainly focus on proper tail behavior.

For modelling joint distribution of two neighboring variables $(d=2)$: using the previous value to predict conditional distribution, the log-likelihood improvement was $\approx 0.0124$ for static model, $\approx 0.0159$ for adaptive. Analogously for three neighboring variables $(d=3)$ the improvement was $\approx 0.0166$ for static model, $\approx 0.0192$ for adaptive.

\section{Adaptive asymmetric EPD}
While EPD is a symmetric distribution, real data might have asymmetric e.g. tail behavior. To include it in parametric model, we can just glue two (\ref{EPDrho}) formulas into asymmetric EPD (AEPD~\cite{aepd}) by using different $\kappa$ shape parameter and/or $\sigma$ scale parameter for the left and right part - generally:
\be \rho(x)=\begin{cases}
      \alpha \frac{C(\kappa_l)}{\sigma_l} e^{-\frac{1}{\kappa_l}\left(\frac{\mu-x}{\sigma_l}\right)^{\kappa_l}}
       & \text{if}\ x<\mu \\
      (1-\alpha) \frac{C(\kappa_r)}{\sigma_r} e^{-\frac{1}{\kappa_r}\left(\frac{x-\mu}{\sigma_r}\right)^{\kappa_r}} & \text{if}\ x\geq \mu
    \end{cases}\label{GAEPG}\ee
for $C(\kappa)=\kappa^{-1/\kappa}/\Gamma(1+1/\kappa)$ normalization as in (\ref{EPDrho}) and $\alpha\in(0,1)$ is probability of the left part ($x<\mu$), $\alpha=1/2$ for standard symmetric.

For $x=\mu$ this parametrization is not necessarily continuous. It often can be ignored, e.g. when using CDF to normalize variable as in \ref{HCR}. If it is an issue, we could smoothen transition e.g. by multiplying the left part by some sigmoid function of $x-\mu$, the right one by one minus this functions, but it would require many arbitrary choices.

Alternatively, we can ensure continuity by satisfying
$$\alpha \frac{C(\kappa_l)}{\sigma_l} = (1-\alpha) \frac{C(\kappa_r)}{\sigma_r}\quad\textrm{condition,}$$
for example by choosing
\be\alpha =\left(\frac{C(\kappa_l)\,\sigma_r}{C(\kappa_r)\,\sigma_l}+1 \right)^{-1}.\label{alpha}\ee

There are many ways for AEPD adaptive estimation, some remarks:
\begin{itemize}
   \item Directly use e.g. $l_T=\sum_{t<T} \eta^{T-t} \ln(\rho_\theta (x_t))$ moving estimator, but it would have high computational cost.
  \item As previously, not optimal but a natural choice for $\mu$ estimator is $\hat{\mu}_{T+1}= (1-\nu) \hat{\mu}_{T} +\nu\, x_T$.
  \item While being only an approximation, it is tempting to treat $x<\mu$ and $x\geq \mu$ as being  correspondingly from left or right separate distribution, updating e.g. scale parameter for exactly one of them:
  $$(\sigma_{l \textrm{ or } r})^\kappa\leftarrow\eta(\sigma_{l \textrm{ or } r})^\kappa +(1-\eta)|x_T-\mu|^\kappa$$
  \item As $\alpha$ corresponds to probability of $x<\mu$, we could update it e.g. as $\alpha\leftarrow \xi \alpha +(1-\xi)[x< \mu]$ for some $\xi\in(0,1)$, where $[c]=1$ when $c$ is true, $0$ otherwise. However, it might be safer to use (\ref{alpha}) continuity condition instead.
  \item We can always use evolution of parameters based on gradients to improve log-likelihood, what should be considered separately for each parameter, using separate (tiny) learning rates $\epsilon >0$ and update e.g.
      $$\theta_{T+1} = \theta_T +\epsilon \left(\frac{\partial}{\partial_\theta} \ln(\rho_{\theta})\right)(x_T)$$
      This is first order method, there might be also considered second order - trying to locally model the evaluation criterion as parabola or paraboloid of parameters and remain in its extremum, briefly discussed in~\cite{adapt}.
\end{itemize}
To summarize, while we could always use moving estimator at high computational cost, choosing a more practical approximation has often large a freedom, also its optimization might be data dependent.

However, such e.g. AEPD model can/should be complemented with further models, like using its CDF to normalize variables to nearly uniform marginal distributions, then e.g. model joint distribution in a window as a polynomial (static or adaptive) as in \ref{HCR}. Such polynomial model can extract and exploit complex behavior of body of the distribution, but not of the tail. Hence on e.g. AEPD normalization level we should mainly focus on getting proper tail behavior, maybe even using general non-continuous (\ref{GAEPG}) form, as continuity is not required for normalization, and this non-continuity can be further smoothed with polynomials.

\section{Conclusions and further work}
While applied time series analysis often uses static Gaussian distribution, there was shown that simple inexpensive generalization: to adaptive distribution, and to more general exponential power distribution, can essentially improve standard evaluation: log-likelihood.\\

This article is focused only on the basic general approaches, in practice it can be further improved by combining with complementing methods, for example mentioned in Section \ref{HCR} additional high parameter modelling of joint distribution for variables normalized with CDF of distributions discussed here.

While discussed adaptive estimation of scale parameter $\sigma$ is MLE-optimal, the remaining parameters rather require some approximations - worth further exploration of better approaches.

Another open question is finding better ways for choosing rate of exponential moving average, also varying in time to include changing rate of forgetting e.g. due to varying time differences.

In contrast e.g. to Levy/stable distributions, the discussed EPD does not cover heavy tails (1/polynomial density) - it is worth to search for practical adaptive estimation also for other types of parametric distributions.

\appendix
This appendix contains Wolfram Mathematica source for used evaluation of adaptive exponential power distribution (vectorized for performance). The \verb"Prepend" inserts initial values in the beginning, then \verb"[[1;;-2]]" removes the last value, so the used density parameter is modeled based only on history:

\begin{small}
\begin{verbatim}
(* xt: sequence of values, kap: fixed kappa *)
(* eta, nu: EMA coefficients *)
(* mu1, sigma1: initial mu, sigma *)
cons = kap^(-1/kap)/2 /Gamma[1 + 1/kap];
mu = ExponentialMovingAverage[
  Prepend[xt, mu1], nu][[1 ;; -2]];
sigma = ExponentialMovingAverage[
  Prepend[Abs[xt - mu]^kap, sigma1^kap]
  , eta][[1 ;; -2]]^(1/kap);
rho=cons*Exp[-((Abs[xt-mu]/sigma)^kap)/kap]/sigma;
Mean[Log[rho]]     (* mean log-likelihood *)
\end{verbatim}
\end{small}

\bibliographystyle{IEEEtran}
\bibliography{cites1}
\end{document}